\documentclass[runningheads]{llncs}

\usepackage[utf8]{inputenc}
\usepackage[T1]{fontenc}
\usepackage[english]{babel}

\usepackage{amsmath}
\usepackage{amsfonts}
\usepackage{amssymb}

\usepackage{graphicx}
\usepackage{float}
\usepackage{subcaption}

\usepackage{algorithm}
\usepackage{algpseudocode}

\usepackage[colorinlistoftodos,prependcaption,textsize=tiny]{todonotes}

\usepackage{hyperref}
\usepackage{url}

\usepackage{longtable, multirow, booktabs, bm}

\DeclareMathOperator*{\argmax}{arg\,max}
\DeclareMathOperator{\clip}{clip}

\title{Large-scale Testing Global Optimization Methods with Black-box Adversarial Attacks}
\titlerunning{Testing global optimization methods with adversarial attacks.}

\author{Wojciech Zarzecki
\and
Jaros{\l}aw Arabas
}
\authorrunning{W. Zarzecki and J. Arabas}
\institute{Warsaw University of Technology, Institute of Computer Science \\
\email{wojciech.zarzecki.stud@pw.edu.pl}, \email{jaroslaw.arabas@pw.edu.pl}}

\begin{document}

\maketitle

\begin{abstract}
Existing global optimization benchmark suites are of a moderate size and are based on a small number of analytical functions that date back even to the 1970s. This causes a risk of biasing the development of global optimization methods. We argue that the tasks related to the black-box adversarial attack (BBAA) can serve as valuable global optimization benchmark in many-dimensional space. We demonstrate the efficiency of several types of evolutionary algorithms and other metaheuristics in solving example BBAA problems.
Thus, we take a step towards convergence of global optimization methods to the challenges and needs that arise in the modern machine learning field.

\end{abstract}

\section{Introduction}

\subsection{Evolution of global optimization benchmarking methods}
Development of global optimization methods is facilitated by benchmark suites. Experimental analysis of global optimization methods dates back at least to the 1970's, with the book by Dixon and Szeg{\"o} \cite{dixon1978tgo2}  being perhaps the most commonly known. The original idea of the authors was to define the test objective functions that are easy to compute, with the known position and value of the global optimum. At the same time, the landscape of the test functions was designed to model the difficulties that were expected to be observed in real-world optimization problems. 

The advent of evolutionary computation has broadened the pool of test functions. As a result, there are several famous test functions whose origins date back to the 1980's or earlier, including functions by Ackley, Griewank, Rastrigin, Rosenbrock, Schaffer, and Schwefel (the names are given to honour their inventors), to name a few.

The year 2005 brought the inspiring competition at the CEC conference that was based on the CEC'2005 benchmarking suite \cite{cec2005} that comprised 25 optimization problems. The number of dimensions was 10, 30, and 50. The benchmark suite came with the computer code that implemented the optimization problems to be run by the author of the optimization method. The benchmark suite came with the standardized testing methodology definition. Subsequent years brought a series of CEC benchmark suites \cite{cec2013,cec2014,cec2015,cec2017,cec2019,cec2020,cec2022}, with the dimension number not exceeding 100.

Another advance towards the reproducibility of results was the COCO suite \cite{hansen2009real1,hansen2009real}. The benchmarking process was performed by the COCO software, and the authors of the optimization method had to implement it as a function to be run by the benchmarking procedure. The number of dimensions did not exceed 40.

In CEC and COCO, the benchmarking can be performed many times for the same method, so the optimization method can be tuned for maximum performance for the whole suite. To avoid this, the BBComp competition \cite{bbcomp} provided access to the server that allowed the user a maximum number of objective function evaluations. In this way, the organizers planned to avoid the risk of ``overfitting'' the optimization methods to the benchmarking suite. In this competition, the maximum number of dimensions was 64.
%
Large-scale versions of the CEC suite \cite{cec2012lsgo,cec2018lsgo} and of  COCO \cite{elhara2019coco} were also proposed, where the dimension number was at most 1000 and 640, respectively. 

In the benchmarking suites from the CEC and COCO families, the optimization problems were defined with the use of 
 combinations of a few types of objective functions that were introduced in much earlier literature. 
In contrast to these 
``synthetic'' problems, the CEC real-world optimization suite \cite{cec2020realworld} was based on problems taken directly from practical applications. The maximum number of dimensions was only 30.

One possible application of optimization methods is the field of Machine Learning (ML). In ML, in particular in the training process of neural networks, standard procedures are gradient-based, since the gradient and the value of the loss function can be computed at comparable cost. The number of dimensions in typical neural models is enormous. Therefore, the ML optimization problems, where the global optimization methods can be considered, are rather related to tuning hyperparameters of the learning process - cf. the black-box optimization challenge at NeurIPS'2020 \cite{turner2021bayesian}, where the number of parameters to optimize is rather moderate.

\subsection{Adversarial attack problem}
\label{sec:related}

\begin{figure}[t!]
    \centering
    \includegraphics[width=1\textwidth]{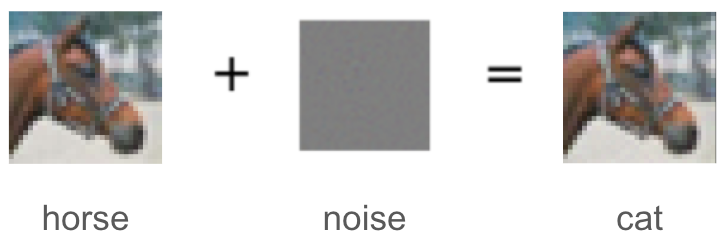}
    \caption{Example adversarial attack: an image that is originally classified as a horse, after adding the noise, is classified as a cat}
    \label{fig:example_attack}
\end{figure}

Progress in machine learning and computer vision has resulted in a broader application of machine learning models, which has naturally attracted attention to the safety and robustness of these mechanisms. One of the key aspects in this area is the adversarial attack. Its goal is a perturbation of an input image that is imperceptible to a human observer but changes the decision of the model. 
Figure~\ref{fig:example_attack} illustrates an example successful black-box attack where a subtle, optimized perturbation is added to an original image of a horse. 
As illustrated in Figure~\ref{fig:example_attack}, the input remains benign to the human eye while causing a wrong decision of the attacked model~\cite{szegedy2014intriguing}. 

Such attacks are worth attention for two main reasons. First, they reveal the scope of possible threats against vision systems, ranging from physical-world attacks against visual classifiers~\cite{eykholt2018robust} to attacks targeting medical deep learning systems~\cite{finlayson2018adversarial}. Second, they have potential for benign applications such as preserving private or copyrighted data, for example, by watermarking deep neural networks through backdooring~\cite{adi2018turning} or by protecting images from malicious diffusion-based editing~\cite{choi2024diffusionguard}.

Adversarial attacks can be distinguished into model-informed and model-agnostic approaches. The model-informed types of attack make use of the information about the disturbance-related sensitivity of the classifier, e.g., the gradient of the loss function \cite{goodfellow2014explaining}, \cite{madry2017towards}, \cite{Carlini2016TowardsET}. Following Costa~et.~al\cite{costa2024deep}, we call attacks from the first group {\em white-box attacks} and from the latter --- {\em black-box attacks}.

One of the approaches for black-box attacks is gradient estimation, represented by \cite{chen2017zoo}, where a coordinate-wise finite-difference gradient estimation applied to the objective from~\cite{Carlini2016TowardsET} is used to optimize the disturbance pattern. 
This result was improved by the GenAttack method~\cite{alzantot2019genattack}, which used a fitness-proportionate selection GA on class-probability feedback.
AutoZOOM~\cite{tu2019autozoom} is another improvement to ZOO that reduces the number of queries needed to perform the attack by  $93\%$ thanks to the use of the autoencoder-compressed perturbation space.
 Ilyas~et~al.~\cite{ilyas2018blackbox} use the NES gradient estimation and demonstrate achieving 2--3 orders of magnitude fewer queries in comparison to the pixel-wise finite differences. 

Some methods try to rely only on the final decision of the model, eg. the chosen class, without accessing the output probability distribution across all classes. One of them is Boundary Attack~\cite{brendel2018boundary} that consists in performing the random walk along the decision boundary starting from a large adversarial perturbation obtained via progressively added Gaussian noise. This approach was refined in \cite{chen2020hopskipjump}, where the authors perform the Monte Carlo boundary gradient estimation and couple it with the binary search. Other methods~\cite{papernot2017practical,ilyas2019priors} rely on the assumption that the distributions of training data are known.   Liu~et.~al\cite{liu2016delving} suggested sharing the adversarial perturbations for models that were trained using data from the same distribution.

However, the most prominent direction to perform the black-box attack is the application of some optimization methods to find perturbations that change the decision of the attacked model. The Square Attack~\cite{andriushchenko2020square} used a random search to find adversarial perturbations based on the fitness function constructed with the model's output probability distribution. This idea was further developed as  SimBA~\cite{guo2019simba}, where the procedure basis is changed from the Cartesian to the discrete cosine.
Other methods use more sophisticated metaheuristics. In PSO Attack~\cite{pso2020blackbox}, the  Particle Swarm Optimization searches the perturbation space, and the authors report a 99.6\% success rate on CIFAR-10 with fewer queries than ZOO. EvoBA~\cite{ilie2021evoba} is an application of an Evolution Strategy to minimize $L_0$ via sparse pixel-level mutations. Some methods limit the perturbation to one pixel only; for example, One Pixel~\cite{su2019onepixel} is an application of Differential Evolution to find a single modified pixel sufficient to flip the class. The efficiency of Genetic Algorithms, Differential Evolution, and CMA-ES for the one-pixel attacks on CIFAR-10 was compared by Clare~et~al.~\cite{clare2024comparative}. Zheng~et.~al~\cite{zheng2025blackboxbenchcomprehensivebenchmarkblackbox} proposed a benchmark for different black-box adversarial attacks, comparing various attacks techniques; however, the impact of the chosen optimization method for score-based attacks remains under-explored.

\subsection{Scope of the paper}
We focus on the Black-Box Adversarial Attack (BBAA), in which only the model output probability distribution can be accessed. We allow for the perturbation of all pixels of the original image.
%
We define the BBAA as an optimization problem that is aimed at finding the perturbation of the original image that results in misclassification and is similar to the noise with the smallest possible variance. We formulate the local search method to demonstrate that the optimization problem is multimodal. For this reason, the BBAA problem can be considered a benchmark to test global optimization methods. We provide an efficiency comparison of several example metaheuristic methods, including the classical Evolutionary Algorithm, four versions of Differential Evolution, Grey Wolf Optimizer, and the INFO (Efficient Optimizer based on Weighted Mean of Vectors) method.

Furthermore, we give possibility to test new methods, proving a framework at \url{https://anonymous.4open.science/r/black-box-adversarial-attacks/} 

\noindent Under the same address, the reader can find detailed information that would exceed the page limit of this text. 

\section{Optimization task related to the adversarial attack problem}

We consider the classification function $c:[0,1]^n \rightarrow \{1,\ldots,k\}$
and the model $p:[0,1]^n \times \mathbb{R}^N \rightarrow P(\{1,...,k\})$, where $P(\{1,...,k\})$ is the space of probability distributions over the space of indices 1,...,$k$. We
assume that the model assumes $N$ tunable real parameters. Given an input
vector $\mathbf{x}$, the model assigns each class $j$ the probability $p_j(\mathbf{x},\theta)$. The class whose probability is highest is considered the winning class that is yielded by the model:

\[
    \hat{c}(\mathbf{x},\theta) = \argmax_{j=1,\ldots,k}\; p_j(\mathbf{x,\theta});
\]
While performing BBAA, we assume that the model parameters have been tuned to maximize the classification accuracy.

The adversarial attack to the input vector $\mathbf{x}$ consists of defining a disturbance vector $\boldsymbol{\delta}(\mathbf{x}) \in [-\varepsilon,\varepsilon]^n$, where $\varepsilon >0 $ is a hyperparameter controlling the maximum perturbation magnitude. The disturbance is added to the original input, and it is expected that  the classifier's output becomes improper:
\[
    \hat{c}\!\left(\clip(\mathbf{x} + \boldsymbol{\delta}),\theta \right)
    \neq c(\mathbf{x})
\]

where

\[
    \clip(\mathbf{y}) =
    \begin{cases}
        0 & y < 0 \\
        y & 0 \leq y \leq 1 \\
        1 & y > 1
    \end{cases}
\]

For the notation brevity, in further text we simply write  $p({\mathbf x})$ and $ \boldsymbol{\delta}$ instead of  $p({\mathbf x},\theta)$ and $\boldsymbol{\delta}(\mathbf{x})$.

The adversarial attack should be hard to recognize for the human --- it is desired that the degree of disturbance is smallest. We model this by the minimization of the squared $L_2$ norm of $\boldsymbol \delta$. Hence, the adversarial attack for a particular image $\mathbf{x}$ is formulated as a bound-constrained minimization problem with the objective function defined as:
\begin{equation}
    \mathcal{L}(\boldsymbol{\delta}
    ) =
    -\log p_{c(\mathbf{x})}\!\left(\mathbf{x} + \boldsymbol{\delta}\right)
    - \alpha \|\boldsymbol{\delta}\|_2^2
    \label{eq:objfun}
\end{equation}

where

    $c(\mathbf{x})$ is the ground-truth class and 

    $\alpha$ is a hyperparameter to control the importance of the $L_2$
          norm of the perturbation.

 The feasible set is a hypercube $[-\varepsilon,\varepsilon]^n$.

The optimization operates in continuous space: the perturbation vector is generated by the optimizer, added element-wise to the original normalized image, and the perturbed image is clipped to the range $[0, 1]$ to ensure valid pixel values.

The attack for each image is treated as a separate optimization problem. The adversarial attack is performed only for these input vectors that are properly classified when no disturbance is added.

\section{Is adversarial attack a global optimization problem?}

If the classification model is nonlinear with respect to {\bf x}, it can be expected that the objective function \eqref{eq:objfun} is multimodal. In this section we validate this claim.

\subsection{Datasets and models}
The experiments are conducted on two standard image classification benchmarks, CIFAR-10~\cite{krizhevsky2009learning} and ImageNet~\cite{russakovsky2015imagenet}. In their raw form, images are represented as tensors with pixel intensities in the range $[0, 255]$. During preprocessing, the images are converted to floating-point tensors normalized to the range $[0, 1]$. CIFAR-10 images are three-channel (RGB) tensors of shape $3 \times 32 \times 32$. ImageNet images are in higher resolution and vary in size; we follow the authors' recommended procedure before passing them to the attacked classification model. First, we resize input images to a tensor of shape $256  \times 256 \times 3$ and then crop the central pixel to a tensor of size $224  \times 224 \times 3$ --- such a tensor is also the input for the adversarial attack procedure.

Before conducting the adversarial attacks, we trained two classification models to serve as the black-box attack targets. When performing an attack, we access only the predicted probabilities for the target classes. For ImageNet, the model was a standard pre-trained Resnet-18~\cite{He2015DeepRL}, while for CIFAR-10, we trained a model based on standard convolutional architectures, similar to VGG~\cite{simonyan2015deepconvolutionalnetworkslargescale}. To ensure reproducibility, we provide complete information about the architectural and training details in the linked archive.

\subsection{Local search applied for the adversarial attack}

We define a local optimization method, the \textbf{Stochastic Growth Attack with Binary Search Refinement} (SGA-BSR), designed to identify a minimal perturbation $\boldsymbol{\delta}$ through a sequential two-phase process. The images are represented in the pixel space as flattened vectors of size $3 \times 32 \times 32$ (CIFAR-10) or $224  \times 224 \times 3$ (ImageNet).

As described in Algorithm~\ref{alg:stochastic_growth}, the SGA-BSR first initiates a \textbf{Stochastic Growth} phase. 
The perturbation vector $\boldsymbol{\delta}$ is initialized to zero. During each of the $N$ iterations, the algorithm randomly selects a single pixel index $idx$ and increments its value by a step size $\eta$, while ensuring the value remains within the valid pixel range $[0, \varepsilon]$ via a clipping function. This modification is accepted only if it results in a non-decreasing objective function value $\mathcal{L}(\boldsymbol{\delta}_{tmp}) \ge \mathcal{L}(\boldsymbol{\delta})$. This greedy, stochastic progression continues until the model $p_\theta$ fails to correctly identify the ground-truth class $c(\mathbf {x})$, at which point the growth phase terminates.

Once a successful misclassification has been achieved, the algorithm enters the \textbf{Refinement} phase 
to minimize the magnitude of the disturbance while maintaining the misclassification. The algorithm identifies the set $S$ of all indices that were modified during the first phase. For each modified index in $S$, a binary search is performed between zero and the current perturbation value $\boldsymbol{\delta}[idx]$. The search seeks to find the smallest possible increment $best\_val$ that still results in an incorrect classification. 
By isolating and minimizing the contribution of each perturbed pixel in sequence, the SGA-BSR effectively identifies a local optimum that satisfies the misclassification constraint with a significantly reduced total perturbation magnitude.
\begin{algorithm}[tbh]
\caption{Stochastic Growth Attack with Binary Search Refinement}
\label{alg:stochastic_growth}
\begin{algorithmic}[1]
\State \textbf{Input:} Original image $\mathbf{x}$, ground-truth $c(\mathbf{x})$, model
    $p_j$, iterations $K$, step size $\eta$, limit $\varepsilon$=1
\State \textbf{Output:} Adversarial image $\mathbf{x}_{adv}$
\State Initialize perturbation $\boldsymbol{\delta} \gets \mathbf{0}$

\State \textbf{Phase 1: Stochastic Growth}
\For{$i = 1$ \textbf{to} $K$}
    \State Choose a random index $idx \in \{1, \dots, C \times H \times W\}$
    \State $\boldsymbol{\delta}_{tmp} \gets \boldsymbol{\delta}$
    \State $\boldsymbol{\delta}_{tmp}[idx] \gets
        \clip(\boldsymbol{\delta}[idx] + \eta,\; 0,\; \varepsilon)$

    \If{$\mathcal{L}(\boldsymbol{\delta}_{tmp}) \ge
        \mathcal{L}(\boldsymbol{\delta})$}
        \State $\boldsymbol{\delta} \gets \boldsymbol{\delta}_{tmp}$
    \EndIf

    \If{$\hat{c}(\text{clip}(\mathbf{x} + \boldsymbol{\delta})) \neq c(\mathbf{x})$}
        \State \textbf{break} \Comment{Target misclassified}
    \EndIf
\EndFor

\State \textbf{Phase 2: Refinement}
\State Identify indices $S = \{idx \mid \boldsymbol{\delta}[idx] > 0\}$

\For{\textbf{each} $idx \in S$}
    \State $low \gets 0, \quad high \gets \boldsymbol{\delta}[idx],
        \quad best\_val \gets high$

    \While{$low \le high$}
        \State $mid \gets \lfloor (low + high) / 2 \rfloor$
        \State $\boldsymbol{\delta}_{test} \gets \boldsymbol{\delta}$
        \State $\boldsymbol{\delta}_{test}[idx] \gets mid$

        \If{$\hat{c}(\text{clip}(\mathbf{x} + \boldsymbol{\delta}_{test})) \neq c(\mathbf{x})$}
            \State $best\_val \gets mid, \quad high \gets mid - 1$
        \Else
            \State $low \gets mid + 1$
        \EndIf
    \EndWhile
    \State $\boldsymbol{\delta}[idx] \gets best\_val$
\EndFor

\State \Return $\mathbf{x}_{adv} \gets \mathbf{x} + \boldsymbol{\delta}$
\end{algorithmic}
\end{algorithm}

We execute SGA-BSR on each image using 1000 different random seeds. If the underlying problem were strictly unimodal (a simple local optimization task), attacks originating from different seeds would consistently converge to the exact same optimal perturbation. However, our empirical results demonstrate that the resulting perturbations, as well as their quality, are different between independent optimization runs. 

\begin{figure}[t!]
    \centering
    \includegraphics[width=1\linewidth]{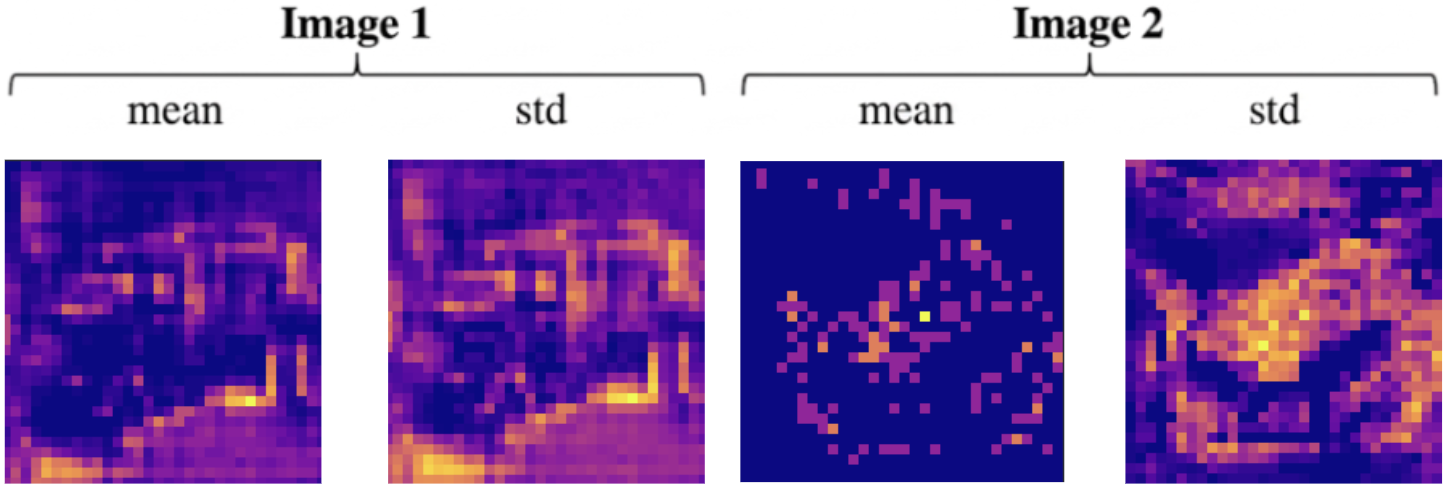}
    \caption{Pixel-wise mean and standard deviation of the perturbation $\boldsymbol{\delta}$ obtained over $100$ independent
    SGA-BSR runs, shown for two example CIFAR-10 images. Regions with a nonzero standard deviation indicate pixels that are used only by some runs. This is evidence that different seeds yield different successful perturbation vectors; therefore, the attack problem is multimodal.}
    \label{fig:heatmaps}
\end{figure}

Figure~\ref{fig:heatmaps} presents heatmaps of the mean and standard deviation of perturbations yielded by each run of Algorithm~\ref{alg:stochastic_growth} for two randomly selected example images. In both cases, it can be observed that for some image regions, the mean disturbance is zero or nearly zero, and the standard deviation is also zero. Yet there are regions of the image for which both the mean and standard deviation of the disturbance are nonzero, which indicates that a diversity of pixels is involved in successful attacks, but there is no need to use all of them to change the model class. In other words, independent runs of SGA-BSR  yield different alternative disturbance vectors $\boldsymbol{\delta}$.  

\begin{figure}[tb]
    \centering
    \includegraphics[width=0.9\linewidth]{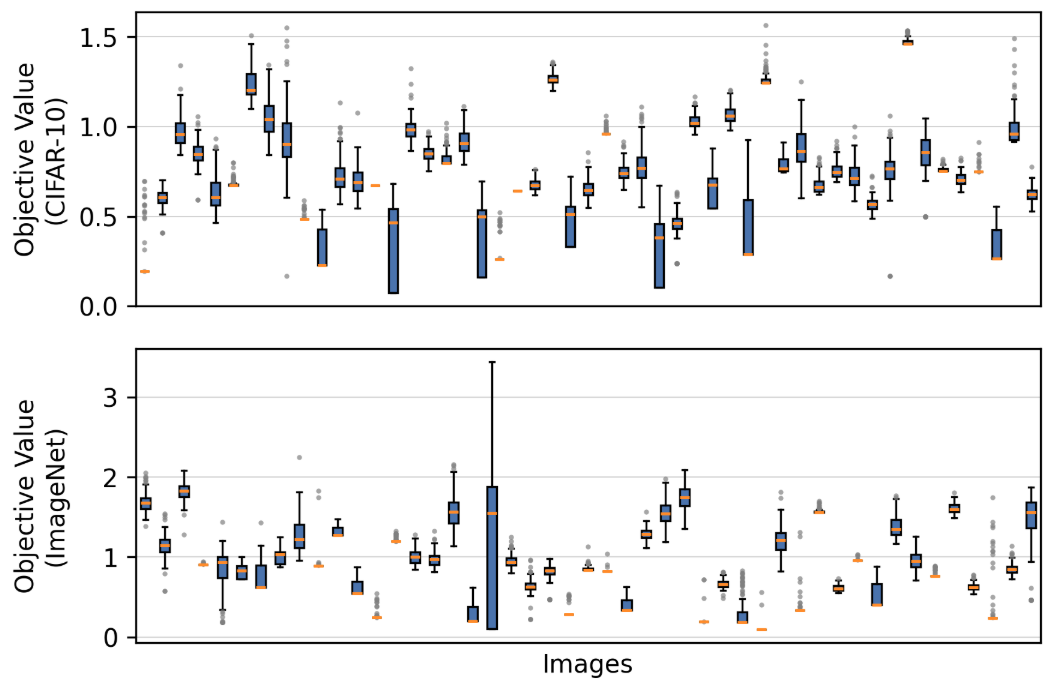}
\caption{\textbf{Per-image variance of SGA-BSR solutions across $1000$
independent runs ($\alpha = 0.1$).} For each image, Algorithm 1 is run with 100 random seeds. Boxplots for objective function value: top two panels: CIFAR-10, bottom: ImageNet. Columns are images, boxes are IQR over seeds, whiskers 1.5 IQR, dots outliers.}
\label{fig:stochastic_growth_box_plot}
    \label{fig:stochastic_growth_box_plot_2}
\end{figure}

Figure~\ref{fig:stochastic_growth_box_plot_2} displays the boxplots of the objective function values across the runs for each image under attack, for CIFAR-10 and ImageNet sets. For many images, the SGA-BSR yielded a variety of results which differed both in the objective function value and the perturbation magnitude. This evidences that the adversarial attack problem has 
many different local optima, since SGA-BSR is a local optimization method. 

\section{Application of global optimization methods to perform black-box adversarial attack}
Knowing that the BBAA is a global optimization problem, we test several global optimization methods to check their efficiency in performing the attack. Section~\ref{sec:outline} describes the details of experiments, while Section~\ref{sec:results} presents results for CIFAR-10 and for ImageNet.

\vspace{-1em}
\subsection{Outline of the experiment}
\label{sec:outline}

 For each image $\mathbf{x}$, we perform 100 independent runs of the optimization method. Each method has the same population size, and in each iteration can test each member of the population only once. The initial population of perturbations generated randomly within the admissible range of $[-\varepsilon, \varepsilon]^n$. Each run of the optimization process is terminated after exceeding the admissible limit of perturbations (which means that the run was unsuccessful). From each run we record the number of tested perturbations, the attack success rate, and the normalized disturbance strength defined as 
 \begin{equation}
     \text{strength}(\boldsymbol{\delta})=\frac{||\boldsymbol{\delta}||_2}{\sqrt{n}}
     \label{eq:L2normalized}
 \end{equation}
 where $n$ is the vector length.

All optimizers were run with their default hyperparameter values as provided by the \texttt{mealpy} library~\cite{van2023mealpy}.
The two settings shared across every optimizer are the population size $N_\text{pop} = 500$ and the maximum number of iterations $T = 500$.
Optimizer-specific defaults are listed in the archive accompanying the paper. 
SADE, GWO, and INFO require no additional hyperparameters beyond the shared ones, as they perform internal self-adaptation or use fixed algorithmic rules.

\subsection{Results}
\label{sec:results}

Table~\ref{tab:adv_all_alpha} summarizes the results obtained for the CIFAR-10 set. Each row corresponds to one combination of optimizer, regularization weight $\alpha$, and maximum pixel disturbance  $\varepsilon$. We report the fraction of attacks that succeed in flipping the label (Suc. R.), the number of queries until the first successful perturbation averaged over successful runs only (First Succ.), the strength of successful perturbations, and the best objective function value observed in each run. Results are reported in the form of pairs (mean $\pm$ standard deviation). 
For each $(\alpha, \varepsilon)$ block, the best objective value across optimizers is highlighted in bold, and the second best is underlined.

The pixel perturbation strength $\varepsilon$ is the single most decisive factor for the attack success on CIFAR-10: at $\varepsilon = 0.01$ every optimizer is effectively blocked, while at $\varepsilon = 0.1$ and $\varepsilon = 0.2$ most methods flip the majority of images. The regularization weight $\alpha$ has a non-monotonic effect: a moderate value of $\alpha = 0.1$ helps most optimizers by pulling the population toward perturbations that both cross the decision boundary and keep a tighter perturbation strength, and in particular lifts GWO 
\noindent from near-zero success, while pushing $\alpha$ up to $10$ or $100$ gives no further efficiency inrease.

{\footnotesize\setlength{\tabcolsep}{1pt}
\begin{longtable}{lcccccc}
\caption{Results of adversarial attacks on the CIFAR-10-based classifier}
\label{tab:adv_all_alpha} \\

\toprule
\textbf{Optim.} & \textbf{$\alpha$} & $\varepsilon$ & \textbf{Suc. R. ($\uparrow$)} & \textbf{First Succ. ($\downarrow$)} & \textbf{$L_2$} & \textbf{Obj. fn. ($\uparrow$)} \\
\midrule
\endfirsthead

\multicolumn{7}{r}{\textit{continued on next page\ldots}} \\[2pt]
\toprule
\textbf{Optim.} & \textbf{$\alpha$} & $\varepsilon$ & \textbf{Suc. R. ($\uparrow$)} & \textbf{First Succ. ($\downarrow$)} & \textbf{$L_2$} & \textbf{Obj. fn. ($\uparrow$)} \\
\midrule
\endhead

\midrule
\multicolumn{7}{r}{\textit{continued on next page\ldots}} \\
\endfoot

\bottomrule
\endlastfoot

DE    & $0.0$ & 0.01 & 2.60\%  & $1.0 \pm 0.0$   & $0.0071 \pm 0.0002$ & \bm{$0.22 \pm 0.26$} \\
GEN   & $0.0$ & 0.01 & 2.60\%  & $1.0 \pm 0.0$   & $0.0057 \pm 0.0001$ & $0.18 \pm 0.23$ \\
GWO   & $0.0$ & 0.01 & 0.00\%  & $0.0 \pm 0.0$   & $0.0000 \pm 0.0000$ & $0.19 \pm 0.24$ \\
INFO  & $0.0$ & 0.01 & 1.30\%  & $1.0 \pm 0.0$   & $0.0052 \pm 0.0000$ & $0.20 \pm 0.24$ \\
JADE  & $0.0$ & 0.01 & 1.30\%  & $1.0 \pm 0.0$   & $0.0055 \pm 0.0000$ & $0.21 \pm 0.26$ \\
SADE  & $0.0$ & 0.01 & 2.60\%  & $1.0 \pm 0.0$   & $0.0062 \pm 0.0001$ & \underline{$0.21 \pm 0.26$} \\
SHADE & $0.0$ & 0.01 & 2.60\%  & $1.0 \pm 0.0$   & $0.0060 \pm 0.0005$ & $0.21 \pm 0.26$ \\
\midrule
DE    & $0.0$ & 0.1  & 28.57\% & $1.0 \pm 0.0$   & $0.0708 \pm 0.0044$ & \bm{$0.75 \pm 0.95$} \\
GEN   & $0.0$ & 0.1  & 24.68\% & $1.0 \pm 0.0$   & $0.0565 \pm 0.0019$ & $0.41 \pm 0.69$ \\
GWO   & $0.0$ & 0.1  & 1.30\%  & $1.0 \pm 0.0$   & $0.0335 \pm 0.0000$ & $0.21 \pm 0.25$ \\
INFO  & $0.0$ & 0.1  & 25.97\% & $1.0 \pm 0.0$   & $0.0563 \pm 0.0047$ & $0.45 \pm 0.66$ \\
JADE  & $0.0$ & 0.1  & 27.27\% & $1.0 \pm 0.0$   & $0.0614 \pm 0.0050$ & $0.60 \pm 0.75$ \\
SADE  & $0.0$ & 0.1  & 28.57\% & $1.0 \pm 0.0$   & $0.0610 \pm 0.0043$ & \underline{$0.63 \pm 0.78$} \\
SHADE & $0.0$ & 0.1  & 25.97\% & $1.0 \pm 0.0$   & $0.0623 \pm 0.0059$ & $0.59 \pm 0.73$ \\
\midrule
DE    & $0.0$ & 0.2  & 59.74\% & $1.0 \pm 0.0$   & $0.1381 \pm 0.0085$ & \bm{$2.47 \pm 2.60$} \\
GEN   & $0.0$ & 0.2  & 57.14\% & $1.0 \pm 0.0$   & $0.1110 \pm 0.0056$ & $1.69 \pm 2.11$ \\
GWO   & $0.0$ & 0.2  & 10.39\% & $1.0 \pm 0.0$   & $0.0658 \pm 0.0006$ & $0.34 \pm 0.47$ \\
INFO  & $0.0$ & 0.2  & 53.25\% & $1.0 \pm 0.0$   & $0.1107 \pm 0.0078$ & $1.62 \pm 2.05$ \\
JADE  & $0.0$ & 0.2  & 54.55\% & $1.0 \pm 0.0$   & $0.1198 \pm 0.0090$ & $1.88 \pm 2.08$ \\
SADE  & $0.0$ & 0.2  & 55.84\% & $1.0 \pm 0.0$   & $0.1208 \pm 0.0074$ & \underline{$1.95 \pm 2.23$} \\
SHADE & $0.0$ & 0.2  & 55.84\% & $1.0 \pm 0.0$   & $0.1189 \pm 0.0103$ & $1.82 \pm 2.05$ \\
\midrule

DE    & $0.1$ & 0.01 & 3.90\%  & $1.0 \pm 0.0$   & $0.0070 \pm 0.0001$ & $0.18 \pm 0.27$ \\
GEN   & $0.1$ & 0.01 & 2.60\%  & $1.0 \pm 0.0$   & $0.0057 \pm 0.0001$ & $0.15 \pm 0.23$ \\
GWO   & $0.1$ & 0.01 & 0.00\%  & $0.0 \pm 0.0$   & $0.0000 \pm 0.0000$ & $0.17 \pm 0.25$ \\
INFO  & $0.1$ & 0.01 & 1.30\%  & $1.0 \pm 0.0$   & $0.0056 \pm 0.0000$ & \bm{$0.19 \pm 0.24$} \\
JADE  & $0.1$ & 0.01 & 2.60\%  & $1.0 \pm 0.0$   & $0.0059 \pm 0.0004$ & $0.18 \pm 0.26$ \\
SADE  & $0.1$ & 0.01 & 1.30\%  & $1.0 \pm 0.0$   & $0.0059 \pm 0.0000$ & $0.18 \pm 0.26$ \\
SHADE & $0.1$ & 0.01 & 2.60\%  & $1.0 \pm 0.0$   & $0.0057 \pm 0.0003$ & \underline{$0.18 \pm 0.26$} \\
\midrule
DE    & $0.1$ & 0.1  & 59.74\% & $14.9 \pm 24.6$ & $0.0742 \pm 0.0054$ & \bm{$0.67 \pm 1.03$} \\
GEN   & $0.1$ & 0.1  & 61.04\% & $11.2 \pm 29.7$ & $0.0569 \pm 0.0018$ & \underline{$0.59 \pm 0.64$} \\
GWO   & $0.1$ & 0.1  & 32.47\% & $16.4 \pm 25.8$ & $0.0888 \pm 0.0141$ & $0.32 \pm 0.76$ \\
INFO  & $0.1$ & 0.1  & 25.97\% & $1.1 \pm 0.5$   & $0.0556 \pm 0.0017$ & $0.37 \pm 0.55$ \\
JADE  & $0.1$ & 0.1  & 50.65\% & $8.9 \pm 12.6$  & $0.0633 \pm 0.0068$ & $0.50 \pm 0.70$ \\
SADE  & $0.1$ & 0.1  & 37.66\% & $11.8 \pm 30.4$ & $0.0630 \pm 0.0064$ & $0.36 \pm 0.76$ \\
SHADE & $0.1$ & 0.1  & 40.26\% & $6.3 \pm 11.6$  & $0.0606 \pm 0.0056$ & $0.53 \pm 0.62$ \\
\midrule
DE    & $0.1$ & 0.2  & 74.03\% & $5.4 \pm 23.5$  & $0.1432 \pm 0.0092$ & \underline{$2.23 \pm 2.43$} \\
GEN   & $0.1$ & 0.2  & 97.40\% & $4.2 \pm 6.7$   & $0.1115 \pm 0.0050$ & \bm{$2.63 \pm 1.60$} \\
GWO   & $0.1$ & 0.2  & 62.34\% & $9.4 \pm 34.9$  & $0.1468 \pm 0.0475$ & $1.43 \pm 2.13$ \\
INFO  & $0.1$ & 0.2  & 53.25\% & $1.0 \pm 0.0$   & $0.1095 \pm 0.0082$ & $1.22 \pm 1.71$ \\
JADE  & $0.1$ & 0.2  & 81.82\% & $6.2 \pm 21.8$  & $0.1228 \pm 0.0111$ & $1.85 \pm 1.92$ \\
SADE  & $0.1$ & 0.2  & 66.23\% & $4.8 \pm 23.3$  & $0.1265 \pm 0.0118$ & $1.58 \pm 2.17$ \\
SHADE & $0.1$ & 0.2  & 89.61\% & $15.4 \pm 42.1$ & $0.1178 \pm 0.0123$ & $1.93 \pm 1.81$ \\
\midrule

DE    & $1.0$ & 0.01 & 2.60\%  & $1.0 \pm 0.0$   & $0.0068 \pm 0.0000$ & $-0.18 \pm 0.27$ \\
GEN   & $1.0$ & 0.01 & 2.60\%  & $1.0 \pm 0.0$   & $0.0057 \pm 0.0001$ & $-0.10 \pm 0.22$ \\
GWO   & $1.0$ & 0.01 & 0.00\%  & $0.0 \pm 0.0$   & $0.0000 \pm 0.0000$ & $-0.23 \pm 0.39$ \\
INFO  & $1.0$ & 0.01 & 2.60\%  & $1.0 \pm 0.0$   & $0.0055 \pm 0.0001$ & \bm{$0.19 \pm 0.23$} \\
JADE  & $1.0$ & 0.01 & 1.30\%  & $1.0 \pm 0.0$   & $0.0054 \pm 0.0000$ & $-0.13 \pm 0.28$ \\
SADE  & $1.0$ & 0.01 & 2.60\%  & $1.0 \pm 0.0$   & $0.0055 \pm 0.0000$ & $-0.14 \pm 0.28$ \\
SHADE & $1.0$ & 0.01 & 2.60\%  & $1.0 \pm 0.0$   & $0.0059 \pm 0.0005$ & \underline{$-0.06 \pm 0.24$} \\
\midrule
DE    & $1.0$ & 0.1  & 49.35\% & $17.1 \pm 40.7$ & $0.0752 \pm 0.0057$ & $-2.87 \pm 1.69$ \\
GEN   & $1.0$ & 0.1  & 81.82\% & $11.9 \pm 25.1$ & $0.0568 \pm 0.0021$ & \underline{$-0.58 \pm 1.61$} \\
GWO   & $1.0$ & 0.1  & 64.94\% & $13.5 \pm 34.1$ & $0.0790 \pm 0.0221$ & $-3.34 \pm 2.16$ \\
INFO  & $1.0$ & 0.1  & 23.38\% & $1.0 \pm 0.0$   & $0.0535 \pm 0.0029$ & \bm{$-0.29 \pm 0.78$} \\
JADE  & $1.0$ & 0.1  & 53.25\% & $20.0 \pm 37.8$ & $0.0629 \pm 0.0102$ & $-2.19 \pm 1.71$ \\
SADE  & $1.0$ & 0.1  & 33.77\% & $1.8 \pm 2.3$   & $0.0613 \pm 0.0084$ & $-2.91 \pm 1.11$ \\
SHADE & $1.0$ & 0.1  & 57.14\% & $25.5 \pm 43.7$ & $0.0496 \pm 0.0078$ & $-0.89 \pm 1.57$ \\
\midrule
DE    & $1.0$ & 0.2  & 74.03\% & $8.3 \pm 31.0$  & $0.1478 \pm 0.0093$ & $-4.30 \pm 3.79$ \\
GEN   & $1.0$ & 0.2  & 92.21\% & $13.2 \pm 41.0$ & $0.1113 \pm 0.0051$ & \bm{$0.85 \pm 3.24$} \\
GWO   & $1.0$ & 0.2  & 79.22\% & $9.1 \pm 22.4$  & $0.1542 \pm 0.0445$ & $-4.98 \pm 4.43$ \\
INFO  & $1.0$ & 0.2  & 54.55\% & $1.0 \pm 0.0$   & $0.1050 \pm 0.0078$ & $-1.79 \pm 2.27$ \\
JADE  & $1.0$ & 0.2  & 74.03\% & $12.2 \pm 45.5$ & $0.1277 \pm 0.0180$ & $-2.72 \pm 3.69$ \\
SADE  & $1.0$ & 0.2  & 68.83\% & $5.6 \pm 18.8$  & $0.1216 \pm 0.0151$ & $-4.49 \pm 3.05$ \\
SHADE & $1.0$ & 0.2  & 76.62\% & $17.4 \pm 43.0$ & $0.1006 \pm 0.0141$ & \underline{$-0.52 \pm 3.21$} \\
\midrule

DE    & $10.0$ & 0.01 & 3.90\%  & $1.3 \pm 0.5$   & $0.0068 \pm 0.0002$ & $-3.77 \pm 0.27$ \\
GEN   & $10.0$ & 0.01 & 5.19\%  & $5.8 \pm 5.1$   & $0.0053 \pm 0.0001$ & $-2.69 \pm 0.27$ \\
GWO   & $10.0$ & 0.01 & 5.19\%  & $11.5 \pm 11.3$ & $0.0069 \pm 0.0022$ & $-5.22 \pm 0.62$ \\
INFO  & $10.0$ & 0.01 & 1.30\%  & $1.0 \pm 0.0$   & $0.0055 \pm 0.0000$ & \bm{$0.16 \pm 0.36$} \\
JADE  & $10.0$ & 0.01 & 3.90\%  & $7.0 \pm 8.5$   & $0.0053 \pm 0.0005$ & $-3.34 \pm 0.35$ \\
SADE  & $10.0$ & 0.01 & 3.90\%  & $1.0 \pm 0.0$   & $0.0058 \pm 0.0003$ & $-3.44 \pm 0.32$ \\
SHADE & $10.0$ & 0.01 & 3.90\%  & $34.3 \pm 47.1$ & $0.0043 \pm 0.0009$ & \underline{$-2.37 \pm 0.29$} \\
\midrule
DE    & $10.0$ & 0.1  & 49.35\% & $17.9 \pm 41.6$ & $0.0687 \pm 0.0061$ & $-38.66 \pm 2.64$ \\
GEN   & $10.0$ & 0.1  & 51.95\% & $14.8 \pm 27.5$ & $0.0537 \pm 0.0024$ & $-27.99 \pm 1.40$ \\
GWO   & $10.0$ & 0.1  & 66.23\% & $13.7 \pm 24.6$ & $0.0730 \pm 0.0218$ & $-45.51 \pm 12.28$ \\
INFO  & $10.0$ & 0.1  & 24.68\% & $1.0 \pm 0.0$   & $0.0501 \pm 0.0030$ & \bm{$-6.66 \pm 11.85$} \\
JADE  & $10.0$ & 0.1  & 48.05\% & $28.9 \pm 47.7$ & $0.0556 \pm 0.0121$ & $-33.11 \pm 5.37$ \\
SADE  & $10.0$ & 0.1  & 31.17\% & $2.0 \pm 4.4$   & $0.0535 \pm 0.0054$ & $-33.96 \pm 3.72$ \\
SHADE & $10.0$ & 0.1  & 38.96\% & $28.4 \pm 63.5$ & $0.0430 \pm 0.0076$ & \underline{$-24.79 \pm 3.01$} \\
\midrule
DE    & $10.0$ & 0.2  & 72.73\% & $5.8 \pm 23.8$  & $0.1283 \pm 0.0100$ & $-73.99 \pm 5.51$ \\
GEN   & $10.0$ & 0.2  & 75.32\% & $16.1 \pm 53.2$ & $0.1063 \pm 0.0059$ & $-54.33 \pm 3.66$ \\
GWO   & $10.0$ & 0.2  & 77.92\% & $6.1 \pm 8.1$   & $0.1123 \pm 0.0385$ & $-74.33 \pm 27.87$ \\
INFO  & $10.0$ & 0.2  & 54.55\% & $1.1 \pm 0.5$   & $0.0965 \pm 0.0086$ & \bm{$-29.47 \pm 27.23$} \\
JADE  & $10.0$ & 0.2  & 72.73\% & $10.5 \pm 35.6$ & $0.0984 \pm 0.0160$ & $-59.53 \pm 10.71$ \\
SADE  & $10.0$ & 0.2  & 68.83\% & $4.0 \pm 12.6$  & $0.1049 \pm 0.0128$ & $-63.63 \pm 9.24$ \\
SHADE & $10.0$ & 0.2  & 72.73\% & $23.0 \pm 69.7$ & $0.0822 \pm 0.0143$ & \underline{$-47.43 \pm 7.80$} \\
\midrule

DE    & $100.0$ & 0.01 & 3.90\%  & $1.3 \pm 0.5$   & $0.0070 \pm 0.0004$ & $-39.40 \pm 1.01$ \\
GEN   & $100.0$ & 0.01 & 3.90\%  & $46.0 \pm 63.6$ & $0.0052 \pm 0.0001$ & $-28.85 \pm 0.26$ \\
GWO   & $100.0$ & 0.01 & 5.19\%  & $8.8 \pm 5.4$   & $0.0077 \pm 0.0014$ & $-54.46 \pm 3.31$ \\
INFO  & $100.0$ & 0.01 & 2.60\%  & $1.0 \pm 0.0$   & $0.0054 \pm 0.0000$ & \bm{$-0.60 \pm 4.76$} \\
JADE  & $100.0$ & 0.01 & 3.90\%  & $11.7 \pm 13.7$ & $0.0055 \pm 0.0008$ & $-35.38 \pm 1.54$ \\
SADE  & $100.0$ & 0.01 & 2.60\%  & $1.0 \pm 0.0$   & $0.0055 \pm 0.0001$ & $-36.19 \pm 1.63$ \\
SHADE & $100.0$ & 0.01 & 3.90\%  & $54.0 \pm 74.2$ & $0.0050 \pm 0.0010$ & \underline{$-25.71 \pm 1.45$} \\
\midrule
DE    & $100.0$ & 0.1  & 48.05\% & $20.1 \pm 49.8$ & $0.0680 \pm 0.0057$ & $-390.13 \pm 22.44$ \\
GEN   & $100.0$ & 0.1  & 33.77\% & $22.8 \pm 53.7$ & $0.0513 \pm 0.0018$ & $-290.15 \pm 1.61$ \\
GWO   & $100.0$ & 0.1  & 64.94\% & $10.9 \pm 14.1$ & $0.0720 \pm 0.0217$ & $-460.48 \pm 121.11$ \\
INFO  & $100.0$ & 0.1  & 20.78\% & $1.0 \pm 0.0$   & $0.0498 \pm 0.0030$ & \bm{$-58.47 \pm 114.74$} \\
JADE  & $100.0$ & 0.1  & 44.16\% & $23.0 \pm 43.0$ & $0.0548 \pm 0.0113$ & $-334.58 \pm 47.72$ \\
SADE  & $100.0$ & 0.1  & 35.06\% & $1.7 \pm 3.4$   & $0.0540 \pm 0.0055$ & $-344.92 \pm 35.42$ \\
SHADE & $100.0$ & 0.1  & 37.66\% & $25.7 \pm 70.1$ & $0.0444 \pm 0.0072$ & \underline{$-255.27 \pm 25.61$} \\
\midrule
DE    & $100.0$ & 0.2  & 74.03\% & $6.9 \pm 35.2$  & $0.1273 \pm 0.0111$ & $-752.37 \pm 54.10$ \\
GEN   & $100.0$ & 0.2  & 63.64\% & $12.7 \pm 47.6$ & $0.1006 \pm 0.0048$ & $-578.90 \pm 3.73$ \\
GWO   & $100.0$ & 0.2  & 79.22\% & $9.7 \pm 20.0$  & $0.1131 \pm 0.0410$ & $-750.02 \pm 282.10$ \\
INFO  & $100.0$ & 0.2  & 51.95\% & $1.0 \pm 0.0$   & $0.0958 \pm 0.0087$ & \bm{$-285.93 \pm 276.88$} \\
JADE  & $100.0$ & 0.2  & 72.73\% & $8.2 \pm 23.9$  & $0.0993 \pm 0.0168$ & $-608.23 \pm 101.65$ \\
SADE  & $100.0$ & 0.2  & 68.83\% & $3.3 \pm 8.4$   & $0.1047 \pm 0.0129$ & $-641.97 \pm 84.39$ \\
SHADE & $100.0$ & 0.2  & 71.43\% & $10.9 \pm 38.0$ & $0.0843 \pm 0.0139$ & \underline{$-492.17 \pm 68.39$} \\

\end{longtable}
}

Across optimizers, INFO behaves like a greedy local search and plateaus at the lowest success rates, whereas DE, GEN, JADE, and SHADE spend more queries yet discover markedly stronger adversarial directions. The smallest perturbation strength is achieved by GEN and SHADE, while DE and GWO drift toward the corners of the admissible area, probably because they have no self-adaptation of parameters like SADE, JADE, or SHADE.

For ImageNet, the dependence of the results quality on $\varepsilon$ mirrors that of CIFAR-10. Yet, low $\varepsilon$ values make the attack much more difficult than for CIFAR-10, and no optimizer succeeds at $\varepsilon = 0.01$. On the other hand, at $\varepsilon = 0.1$, most attacks are successful, and several optimizers obtain even complete success at $\varepsilon = 0.2$.
The smallest disturbance strength was obtained by GEN, INFO, and SHADE. Assuming $\alpha = 0.1$ helped nearly all optimizers to obtain good results, provided that the budget was large enough. The most striking beneficiary is again GWO, which, without regularization, stalls near a 10-15\% success rate, but then it catches up to the other population-based methods.
Figure~\ref{fig:all_ships} presents examples of successful attacks for an example image.

{\footnotesize\setlength{\tabcolsep}{1pt}
\begin{longtable}{lcccccc}
\caption{Results of adversarial attacks on the ImageNet-based classifier}
\label{tab:adv_all_alpha_imagenet} \\

\toprule
\textbf{Optim.} & \textbf{$\alpha$} & $\varepsilon$ & \textbf{Suc. R. ($\uparrow$)} & \textbf{First Succ. ($\downarrow$)} & \textbf{$L_2$} & \textbf{Obj. fn. ($\uparrow$)} \\
\midrule
\endfirsthead

\multicolumn{7}{r}{\textit{continued on next page\ldots}} \\[2pt]
\toprule
\textbf{Optim.} & \textbf{$\alpha$} & $\varepsilon$ & \textbf{Suc. R. ($\uparrow$)} & \textbf{First Succ. ($\downarrow$)} & \textbf{$L_2$} & \textbf{Obj. fn. ($\uparrow$)} \\
\midrule
\endhead

\midrule
\multicolumn{7}{r}{\textit{continued on next page\ldots}} \\
\endfoot

\bottomrule
\endlastfoot

DE    & $0.0$ & 0.01 & 0.00\%   & $0.0 \pm 0.0$   & $0.0000 \pm 0.0000$ & \bm{$0.34 \pm 0.49$} \\
GEN   & $0.0$ & 0.01 & 0.00\%   & $0.0 \pm 0.0$   & $0.0000 \pm 0.0000$ & $0.21 \pm 0.32$ \\
GWO   & $0.0$ & 0.01 & 0.00\%   & $0.0 \pm 0.0$   & $0.0000 \pm 0.0000$ & $0.25 \pm 0.36$ \\
INFO  & $0.0$ & 0.01 & 0.00\%   & $0.0 \pm 0.0$   & $0.0000 \pm 0.0000$ & $0.25 \pm 0.36$ \\
JADE  & $0.0$ & 0.01 & 0.00\%   & $0.0 \pm 0.0$   & $0.0000 \pm 0.0000$ & $0.32 \pm 0.46$ \\
SADE  & $0.0$ & 0.01 & 0.00\%   & $0.0 \pm 0.0$   & $0.0000 \pm 0.0000$ & $0.32 \pm 0.47$ \\
SHADE & $0.0$ & 0.01 & 0.00\%   & $0.0 \pm 0.0$   & $0.0000 \pm 0.0000$ & \underline{$0.32 \pm 0.47$} \\
\midrule
DE    & $0.0$ & 0.1  & 60.00\%  & $1.0 \pm 0.0$   & $0.0686 \pm 0.0041$ & \bm{$2.26 \pm 2.09$} \\
GEN   & $0.0$ & 0.1  & 55.00\%  & $1.0 \pm 0.0$   & $0.0552 \pm 0.0025$ & $1.58 \pm 1.76$ \\
GWO   & $0.0$ & 0.1  & 10.00\%  & $1.0 \pm 0.0$   & $0.0331 \pm 0.0000$ & $0.46 \pm 0.82$ \\
INFO  & $0.0$ & 0.1  & 50.00\%  & $1.0 \pm 0.0$   & $0.0567 \pm 0.0072$ & $1.49 \pm 1.75$ \\
JADE  & $0.0$ & 0.1  & 55.00\%  & $1.0 \pm 0.0$   & $0.0610 \pm 0.0058$ & $1.82 \pm 1.80$ \\
SADE  & $0.0$ & 0.1  & 55.00\%  & $1.0 \pm 0.0$   & $0.0596 \pm 0.0035$ & \underline{$1.85 \pm 1.81$} \\
SHADE & $0.0$ & 0.1  & 55.00\%  & $1.0 \pm 0.0$   & $0.0585 \pm 0.0072$ & $1.82 \pm 1.84$ \\
\midrule
DE    & $0.0$ & 0.2  & 100.00\% & $1.0 \pm 0.0$   & $0.1325 \pm 0.0095$ & \bm{$5.61 \pm 2.97$} \\
GEN   & $0.0$ & 0.2  & 85.00\%  & $1.0 \pm 0.0$   & $0.1063 \pm 0.0064$ & $4.18 \pm 2.85$ \\
GWO   & $0.0$ & 0.2  & 15.00\%  & $1.0 \pm 0.0$   & $0.0648 \pm 0.0003$ & $1.15 \pm 1.61$ \\
INFO  & $0.0$ & 0.2  & 85.00\%  & $1.1 \pm 0.5$   & $0.1036 \pm 0.0055$ & $4.11 \pm 2.77$ \\
JADE  & $0.0$ & 0.2  & 90.00\%  & $1.0 \pm 0.0$   & $0.1142 \pm 0.0118$ & $4.70 \pm 2.84$ \\
SADE  & $0.0$ & 0.2  & 90.00\%  & $1.0 \pm 0.0$   & $0.1170 \pm 0.0086$ & \underline{$4.74 \pm 2.84$} \\
SHADE & $0.0$ & 0.2  & 90.00\%  & $1.0 \pm 0.0$   & $0.1167 \pm 0.0105$ & $4.59 \pm 2.84$ \\
\midrule

DE    & $0.1$ & 0.01 & 0.00\%   & $0.0 \pm 0.0$    & $0.0000 \pm 0.0000$ & $0.06 \pm 0.49$ \\
GEN   & $0.1$ & 0.01 & 0.00\%   & $0.0 \pm 0.0$    & $0.0000 \pm 0.0000$ & $0.01 \pm 0.30$ \\
GWO   & $0.1$ & 0.01 & 0.00\%   & $0.0 \pm 0.0$    & $0.0000 \pm 0.0000$ & $-0.04 \pm 0.46$ \\
INFO  & $0.1$ & 0.01 & 0.00\%   & $0.0 \pm 0.0$    & $0.0000 \pm 0.0000$ & \bm{$0.25 \pm 0.36$} \\
JADE  & $0.1$ & 0.01 & 0.00\%   & $0.0 \pm 0.0$    & $0.0000 \pm 0.0000$ & $0.08 \pm 0.47$ \\
SADE  & $0.1$ & 0.01 & 0.00\%   & $0.0 \pm 0.0$    & $0.0000 \pm 0.0000$ & $0.07 \pm 0.47$ \\
SHADE & $0.1$ & 0.01 & 0.00\%   & $0.0 \pm 0.0$    & $0.0000 \pm 0.0000$ & \underline{$0.12 \pm 0.44$} \\
\midrule
DE    & $0.1$ & 0.1  & 80.00\%  & $1.7 \pm 1.7$    & $0.0725 \pm 0.0041$ & $0.43 \pm 2.34$ \\
GEN   & $0.1$ & 0.1  & 90.00\%  & $4.5 \pm 8.4$    & $0.0555 \pm 0.0024$ & \bm{$2.02 \pm 2.25$} \\
GWO   & $0.1$ & 0.1  & 90.00\%  & $5.5 \pm 3.1$    & $0.0832 \pm 0.0180$ & $0.46 \pm 2.32$ \\
INFO  & $0.1$ & 0.1  & 50.00\%  & $1.0 \pm 0.0$    & $0.0537 \pm 0.0022$ & $0.37 \pm 1.06$ \\
JADE  & $0.1$ & 0.1  & 80.00\%  & $7.0 \pm 11.5$   & $0.0637 \pm 0.0078$ & $0.63 \pm 2.02$ \\
SADE  & $0.1$ & 0.1  & 80.00\%  & $16.5 \pm 50.9$  & $0.0620 \pm 0.0055$ & $0.13 \pm 2.10$ \\
SHADE & $0.1$ & 0.1  & 80.00\%  & $14.2 \pm 27.3$  & $0.0528 \pm 0.0051$ & \underline{$0.79 \pm 1.79$} \\
\midrule
DE    & $0.1$ & 0.2  & 100.00\% & $1.0 \pm 0.0$    & $0.1396 \pm 0.0088$ & $1.32 \pm 2.55$ \\
GEN   & $0.1$ & 0.2  & 100.00\% & $1.4 \pm 1.2$    & $0.1069 \pm 0.0061$ & \bm{$3.90 \pm 2.13$} \\
GWO   & $0.1$ & 0.2  & 100.00\% & $3.6 \pm 1.4$    & $0.1474 \pm 0.0457$ & $1.47 \pm 1.95$ \\
INFO  & $0.1$ & 0.2  & 85.00\%  & $1.0 \pm 0.0$    & $0.1022 \pm 0.0079$ & $0.48 \pm 2.17$ \\
JADE  & $0.1$ & 0.2  & 100.00\% & $2.9 \pm 5.7$    & $0.1229 \pm 0.0181$ & $1.54 \pm 2.16$ \\
SADE  & $0.1$ & 0.2  & 100.00\% & $1.9 \pm 3.7$    & $0.1225 \pm 0.0135$ & $1.02 \pm 2.87$ \\
SHADE & $0.1$ & 0.2  & 95.00\%  & $1.1 \pm 0.4$    & $0.1039 \pm 0.0177$ & \underline{$1.69 \pm 2.24$} \\

\end{longtable}

\begin{figure}[!h]
    \centering
    \begin{subfigure}[t]{0.3\linewidth}
        \includegraphics[width=\linewidth]{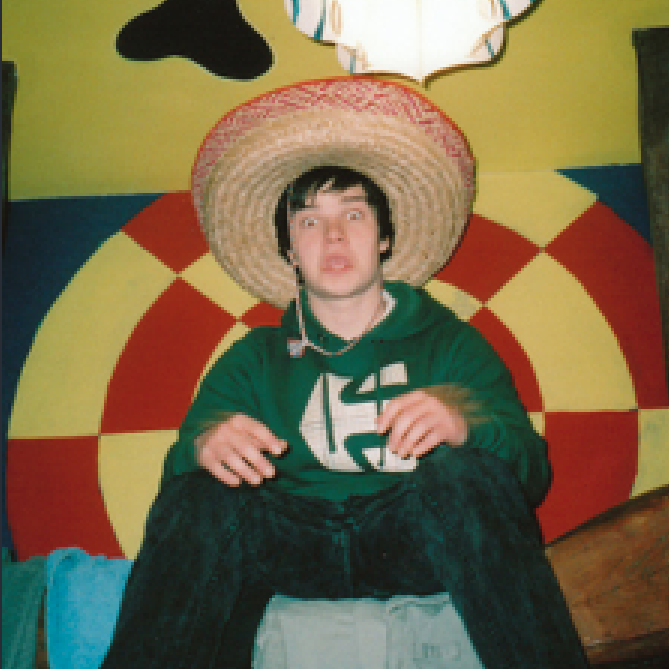}
        \caption{original class: sombrero}
    \end{subfigure}
    \hfill
    \begin{subfigure}[t]{0.3\linewidth}
        \includegraphics[width=\linewidth]{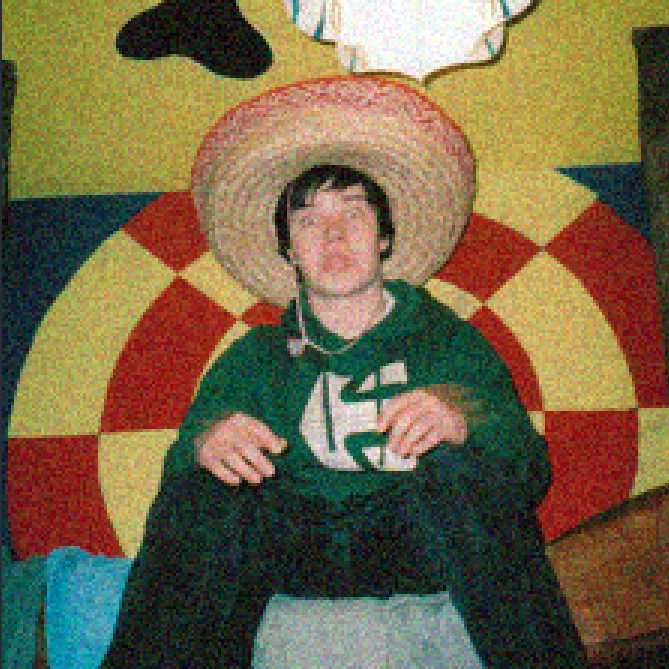}
        \caption{$\varepsilon = 0.1$:  jigsaw puzzle}
    \end{subfigure}
    \hfill
    \begin{subfigure}[t]{0.3\linewidth}
        \includegraphics[width=\linewidth]{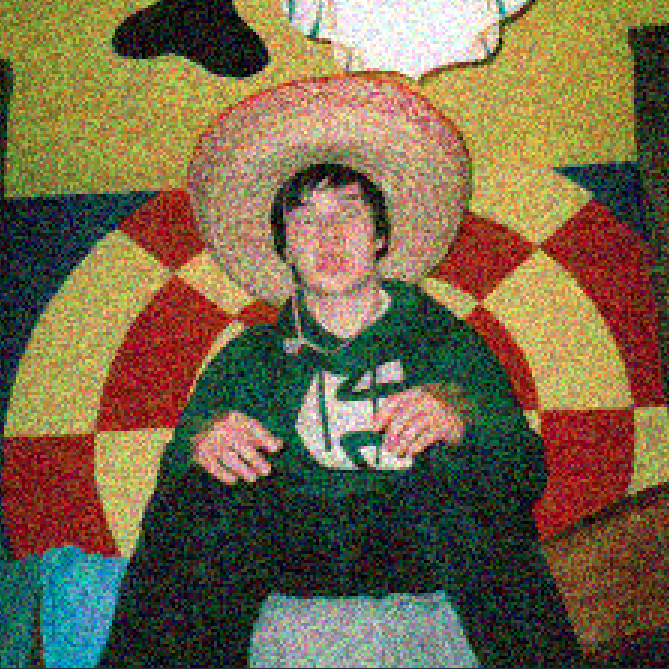}
        \caption{$\varepsilon = 0.2$: jigsaw puzzle}
    \end{subfigure}
    \caption{Example image from ImageNet with two successful attack examples  for different $\varepsilon$ values}
    \label{fig:all_ships}
\end{figure}

At matched $(\varepsilon, \alpha)$, the attack is uniformly easier on ImageNet than on CIFAR-10, which we attribute to the much higher input dimensionality and the much larger label set that statistically shrinks the margin to the nearest competing class. The relative ranking of methods remains stable across datasets: GEN and SHADE dominate by achieving high success rates while keeping low disturbance strength, GWO performs worst whenever the objective lacks a regularization signal, and INFO stays the query-efficient but low-ceiling option.

\section{Discussion and Future Work}
We demonstrated that the BBAA is a demanding global optimization task --- it is multimodal, and the search space has very many dimensions. We provided results of several ready-for-use optimizers, without any parameter tuning. The next step would be to broaden the portfolio of optimization methods and to tune them to achieve better efficiency.
 In future work, we plan to extend the evaluation formula to attacks targeted at falsely assigning a specific class. We also plan to add more classifiers and datasets, and to reformulate the objective function \eqref{eq:objfun} by replacing or complementing the $L_2$ regularization term with a perceptual loss~\cite{Zhang2018TheUE} that better reflects human-visible distortion.

\subsubsection*{Acknowledgment}
{\footnotesize
We gratefully acknowledge Polish high-performance computing infrastructure PLGrid (HPC Center: ACK Cyfronet AGH) for providing computer facilities and support within computational grant no. PLG/2025/018167.
}

\bibliographystyle{splncs04}
\bibliography{reference}

@misc{simonyan2015deepconvolutionalnetworkslargescale,
  title={Very Deep Convolutional Networks for Large-Scale Image Recognition},
  author={Karen Simonyan and Andrew Zisserman},
  journal={CoRR},
  year={2014},
  volume={abs/1409.1556},
  url={https://api.semanticscholar.org/CorpusID:14124313}
}

@inproceedings{chen2017zoo,
  title={ZOO: Zeroth Order Optimization Based Black-box Attacks to Deep Neural Networks without Training Substitute Models},
  author={Pin-Yu Chen and Huan Zhang and Yash Sharma and Jinfeng Yi and Cho-Jui Hsieh},
  journal={Proceedings of the 10th ACM Workshop on Artificial Intelligence and Security},
  year={2017},
  url={https://api.semanticscholar.org/CorpusID:2179389}
}

@inproceedings{ilyas2018blackbox,
  title={Black-box Adversarial Attacks with Limited Queries and Information},
  author={Andrew Ilyas and Logan Engstrom and Anish Athalye and Jessy Lin},
  booktitle={International Conference on Machine Learning},
  year={2018},
  url={https://api.semanticscholar.org/CorpusID:5046541}
}

@inproceedings{tu2019autozoom,
  title={AutoZOOM: Autoencoder-based Zeroth Order Optimization Method for Attacking Black-box Neural Networks},
  author={Chun-Chen Tu and Pai-Shun Ting and Pin-Yu Chen and Sijia Liu and Huan Zhang and Jinfeng Yi and Cho-Jui Hsieh and Shin-Ming Cheng},
  booktitle={AAAI Conference on Artificial Intelligence},
  year={2018},
  url={https://api.semanticscholar.org/CorpusID:44079102}
}

@inproceedings{guo2019simba,
  title={Simple Black-box Adversarial Attacks},
  author={Chuan Guo and Jacob R. Gardner and Yurong You and Andrew Gordon Wilson and Kilian Q. Weinberger},
  journal={ArXiv},
  year={2019},
  volume={abs/1905.07121},
  url={https://api.semanticscholar.org/CorpusID:86541092}
}

@inproceedings{andriushchenko2020square,
  title={Square attack: a query-efficient black-box adversarial attack via random search},
  author={Andriushchenko, Maksym and Croce, Francesco and Flammarion, Nicolas and Hein, Matthias},
  booktitle={European conference on computer vision},
  pages={484--501},
  year={2020},
  organization={Springer}
}

@inproceedings{brendel2018boundary,
  title={Decision-based adversarial attacks: Reliable attacks against black-box machine learning models},
  author={Brendel, Wieland and Rauber, Jonas and Bethge, Matthias},
  journal={arXiv preprint arXiv:1712.04248},
  year={2017}
}

@inproceedings{chen2020hopskipjump,
  title={Hopskipjumpattack: A query-efficient decision-based attack},
  author={Chen, Jianbo and Jordan, Michael I and Wainwright, Martin J},
  booktitle={2020 ieee symposium on security and privacy (sp)},
  pages={1277--1294},
  year={2020},
  organization={IEEE}
}

@inproceedings{alzantot2019genattack,
  title={Genattack: Practical black-box attacks with gradient-free optimization},
  author={Alzantot, Moustafa and Sharma, Yash and Chakraborty, Supriyo and Zhang, Huan and Hsieh, Cho-Jui and Srivastava, Mani B},
  booktitle={Proceedings of the genetic and evolutionary computation conference},
  pages={1111--1119},
  year={2019}
}

@article{su2019onepixel,
  title={One pixel attack for fooling deep neural networks},
  author={Su, Jiawei and Vargas, Danilo Vasconcellos and Sakurai, Kouichi},
  journal={IEEE Transactions on Evolutionary Computation},
  volume={23},
  number={5},
  pages={828--841},
  year={2019},
  publisher={IEEE}
}

@inproceedings{pso2020blackbox,
  title={They might not be giants: Crafting black-box adversarial examples with fewer queries using particle swarm optimization},
  author={Mosli, Rayan and Wright, Matthew and Yuan, Bo and Pan, Yin},
  journal={arXiv preprint arXiv:1909.07490},
  year={2019}
}

@inproceedings{ilie2021evoba,
  title={Evoba: An evolution strategy as a strong baseline for black-box adversarial attacks},
  author={Ilie, Andrei and Popescu, Marius and Stefanescu, Alin},
  booktitle={International Conference on Neural Information Processing},
  pages={188--200},
  year={2021},
  organization={Springer}
}

@inproceedings{clare2024comparative,
  title={A Comparative Analysis of Evolutionary Adversarial One-Pixel Attacks},
  author={Clare, Luana and Marques, Alexandra and Correia, Jo{\~a}o},
  booktitle={International Conference on the Applications of Evolutionary Computation (Part of EvoStar)},
  pages={147--162},
  year={2024},
  organization={Springer}
}

@inproceedings{ilyas2019priors,
  title={Prior convictions: Black-box adversarial attacks with bandits and priors},
  author={Ilyas, Andrew and Engstrom, Logan and Madry, Aleksander},
  journal={arXiv preprint arXiv:1807.07978},
  year={2018}
}

@book{dixon1978tgo2,
    editor = {L. C. W. Dixon and G. P. Szegő},
    title = {Towards Global Optimisation 2},
    year = {1978},
    publisher = {North-Holland Publishing Company},
    address = {Amsterdam},
    isbn = {9780444851710}
}

@article{elhara2019coco,
  title={{COCO}: the large scale black-box optimization benchmarking (bbob-largescale) test suite},
  author={Elhara, Ouassim and Varelas, Konstantinos and Nguyen, Duc and Tusar, Tea and Brockhoff, Dimo and Hansen, Nikolaus and Auger, Anne},
  journal={arXiv preprint arXiv:1903.06396},
  year={2019}
}

@techreport{hansen2009real1,
  title={Real-Parameter Black-Box Optimization Benchmarking 2009: Noiseless Functions Definitions},
  author={Nikolaus Hansen and Raymond Ros and Anne Auger},
  year={2009},
  url={https://api.semanticscholar.org/CorpusID:270969128}
}

@techreport{hansen2009real,
  TITLE = {{Real-Parameter Black-Box Optimization Benchmarking 2009: Noisy Functions Definitions}},
  AUTHOR = {Hansen, Nikolaus and Finck, Steffen and Ros, Raymond and Auger, Anne},
  URL = {https://inria.hal.science/inria-00369466},
  TYPE = {Research Report},
  NUMBER = {RR-6869},
  INSTITUTION = {{INRIA}},
  YEAR = {2009},
}

@techreport{cec2012lsgo,
  title        = {Benchmark Functions for the {CEC}'2012 Special Session and Competition on Large Scale Global Optimization},
  author       = {Omidvar, Mohammad Nabi and Li, Xiaodong and Tang, Ke and Mei, Yi and Yao, Xin},
  institution  = {Nanyang Technological University},
  year         = {2012}
}

@techreport{cec2018lsgo,
  title        = {Benchmark Functions for the {CEC}'2018 Competition on Large Scale Global Optimization},
  author       = {Omidvar, Mohammad Nabi and Li, Xiaodong and Tang, Ke and Yao, Xin},
  institution  = {Nanyang Technological University},
  year         = {2018}
}

@inproceedings{turner2021bayesian,
  title={Bayesian optimization is superior to random search for machine learning hyperparameter tuning: Analysis of the black-box optimization challenge 2020},
  author={Turner, Ryan and Eriksson, David and McCourt, Michael and Kiili, Juha and Laaksonen, Eero and Xu, Zhen and Guyon, Isabelle},
  booktitle={NeurIPS 2020 competition and demonstration track},
  pages={3--26},
  year={2021},
  organization={PMLR}
}

@techreport{cec2005,
  title        = {Problem Definitions and Evaluation Criteria for the {CEC}'2005 Special Session on Real-Parameter Optimization},
  author       = {Liang, Jing J. and Suganthan, Ponnuthurai N. and Deb, Kalyanmoy},
  institution  = {Nanyang Technological University},
  year         = {2005}
}

@techreport{cec2013,
  title        = {Problem Definitions and Evaluation Criteria for the {CEC}'2013 Special Session on Real-Parameter Optimization},
  author       = {Mallipeddi, Rammohan and Suganthan, Ponnuthurai N. and Pan, Quanke and Tasgetiren, Mehmet Fatih},
  institution  = {Nanyang Technological University},
  year         = {2013}
}

@techreport{cec2014,
  title        = {Problem Definitions and Evaluation Criteria for the {CEC}'2014 Special Session and Competition on Single Objective Real-Parameter Optimization},
  author       = {Liang, Jing J. and Qu, Bo-Yang and Suganthan, Ponnuthurai N.},
  institution  = {Zhengzhou University and Nanyang Technological University},
  year         = {2013}
}

@inproceedings{cec2015,
   title={Problem definitions and evaluation criteria for {CEC}'2015 special session on bound constrained single-objective computationally expensive numerical optimization},
  author={Chen, Qin and Liu, B and Zhang, Qiang and Liang, Jing and Suganthan, Ponnuthurai and Qu, Boyang},
  journal={Technical Report, Computational Intelligence Laboratory, Zhengzhou University, Zhengzhou, China and Technical Report, Nanyang Technological University},
  year={2014}
}

@techreport{cec2017,
  title        = {Problem Definitions and Evaluation Criteria for the {CEC}'2017 Competition on Single Objective Real-Parameter Optimization},
  author       = {Liang, Jing J. and Qu, Bo-Yang and Suganthan, Ponnuthurai N. and Chen, Qiang},
  institution  = {Nanyang Technological University},
  year         = {2017}
}

@techreport{cec2019,
  title        = {Problem Definitions and Evaluation Criteria for the {CEC}'2019 Competition on Single Objective Real-Parameter Optimization},
  author       = {Liang, Jing J. and Qu, Bo-Yang and Suganthan, Ponnuthurai N.},
  institution  = {Nanyang Technological University},
  year         = {2018}
}

@techreport{cec2020,
  title        = {Problem Definitions and Evaluation Criteria for the {CEC}'2020 Competition on Single Objective Real-Parameter Optimization},
  author       = {Liang, Jing J. and Qu, Bo-Yang and Suganthan, Ponnuthurai N.},
  institution  = {Nanyang Technological University},
  year         = {2020}
}

@techreport{cec2022,
  title        = {Problem Definitions and Evaluation Criteria for the {CEC}'2022 Competition on Single Objective Real-Parameter Optimization},
  author       = {Liang, Jing J. and Qu, Bo-Yang and Suganthan, Ponnuthurai N.},
  institution  = {Nanyang Technological University},
  year         = {2022}
}

@techreport{cec2020realworld,
  title        = {Problem Definitions and Evaluation Criteria for the {CEC}'2011 Competition on Testing Evolutionary Algorithms on Real world Problems},
  author       = {Das, Swagatam and Suganthan, Ponnuthurai N.},
  year         = {2010}
}

@misc{bbcomp,
  title        = {Black-Box Optimization Competition (BBComp)},
  howpublished = {\url{https://www.ini.rub.de/PEOPLE/glasmtbl/projects/bbcomp/index.html}},
  author       = {{Ruhr University Bochum, Institute for Neural Computation}}
}

@inproceedings{szegedy2014intriguing,
  title={Intriguing properties of neural networks},
  author={Szegedy, Christian and Zaremba, Wojciech and Sutskever, Ilya and Bruna, Joan and Erhan, Dumitru and Goodfellow, Ian and Fergus, Rob},
  journal={arXiv preprint arXiv:1312.6199},
  year={2013}
}

@inproceedings{eykholt2018robust,
  title={Robust physical-world attacks on deep learning visual classification},
  author={Eykholt, Kevin and Evtimov, Ivan and Fernandes, Earlence and Li, Bo and Rahmati, Amir and Xiao, Chaowei and Prakash, Atul and Kohno, Tadayoshi and Song, Dawn},
  booktitle={Proceedings of the IEEE conference on computer vision and pattern recognition},
  pages={1625--1634},
  year={2018}
}

@article{finlayson2018adversarial,
  title={Adversarial attacks against medical deep learning systems},
  author={Finlayson, Samuel G and Chung, Hyung Won and Kohane, Isaac S and Beam, Andrew L},
  journal={arXiv preprint arXiv:1804.05296},
  year={2018}
}

@inproceedings{adi2018turning,
  title={Turning your weakness into a strength: Watermarking deep neural networks by backdooring},
  author={Adi, Yossi and Baum, Carsten and Cisse, Moustapha and Pinkas, Benny and Keshet, Joseph},
  booktitle={27th USENIX security symposium (USENIX Security 18)},
  pages={1615--1631},
  year={2018}
}

@inproceedings{choi2024diffusionguard,
  title={Diffusionguard: A robust defense against malicious diffusion-based image editing},
  author={Choi, June Suk and Lee, Kyungmin and Jeong, Jongheon and Xie, Saining and Shin, Jinwoo and Lee, Kimin},
  journal={arXiv preprint arXiv:2410.05694},
  year={2024}
}

@article{goodfellow2014explaining,
  title={Explaining and harnessing adversarial examples},
  author={Goodfellow, Ian J and Shlens, Jonathon and Szegedy, Christian},
  journal={arXiv preprint arXiv:1412.6572},
  year={2014}
}

@article{madry2017towards,
  title={Towards deep learning models resistant to adversarial attacks},
  author={Madry, Aleksander and Makelov, Aleksandar and Schmidt, Ludwig and Tsipras, Dimitris and Vladu, Adrian},
  journal={arXiv preprint arXiv:1706.06083},
  year={2017}
}

@article{Carlini2016TowardsET,
  title={Towards Evaluating the Robustness of Neural Networks},
  author={Nicholas Carlini and David A. Wagner},
  journal={2017 IEEE Symposium on Security and Privacy (SP)},
  year={2016},
  pages={39-57},
  url={https://api.semanticscholar.org/CorpusID:2893830}
}

@article{costa2024deep,
  title={How deep learning sees the world: A survey on adversarial attacks \& defenses},
  author={Costa, Joana C and Roxo, Tiago and Proen{\c{c}}a, Hugo and Inacio, Pedro Ricardo Morais},
  journal={IEEE Access},
  volume={12},
  pages={61113--61136},
  year={2024},
  publisher={IEEE}
}

@article{liu2016delving,
  title={Delving into transferable adversarial examples and black-box attacks},
  author={Liu, Yanpei and Chen, Xinyun and Liu, Chang and Song, Dawn},
  journal={arXiv preprint arXiv:1611.02770},
  year={2016}
}

@inproceedings{papernot2017practical,
  title={Practical black-box attacks against machine learning},
  author={Papernot, Nicolas and McDaniel, Patrick and Goodfellow, Ian and Jha, Somesh and Celik, Z Berkay and Swami, Ananthram},
  booktitle={Proceedings of the 2017 ACM on Asia conference on computer and communications security},
  pages={506--519},
  year={2017}
}

@article{krizhevsky2009learning,
  title={Learning multiple layers of features from tiny images},
  author={Krizhevsky, Alex and Hinton, Geoffrey and others},
  year={2009},
  publisher={Toronto, ON, Canada}
}

@article{russakovsky2015imagenet,
  title={Imagenet large scale visual recognition challenge},
  author={Russakovsky, Olga and Deng, Jia and Su, Hao and Krause, Jonathan and Satheesh, Sanjeev and Ma, Sean and Huang, Zhiheng and Karpathy, Andrej and Khosla, Aditya and Bernstein, Michael and others},
  journal={International journal of computer vision},
  volume={115},
  number={3},
  pages={211--252},
  year={2015},
  publisher={Springer}
}

@article{He2015DeepRL,
  title={Deep Residual Learning for Image Recognition},
  author={Kaiming He and X. Zhang and Shaoqing Ren and Jian Sun},
  journal={2016 IEEE Conference on Computer Vision and Pattern Recognition (CVPR)},
  year={2015},
  pages={770-778},
  url={https://api.semanticscholar.org/CorpusID:206594692}
}

@article{Zhang2018TheUE,
  title={The Unreasonable Effectiveness of Deep Features as a Perceptual Metric},
  author={Richard Zhang and Phillip Isola and Alexei A. Efros and Eli Shechtman and Oliver Wang},
  journal={2018 IEEE/CVF Conference on Computer Vision and Pattern Recognition},
  year={2018},
  pages={586-595},
  url={https://api.semanticscholar.org/CorpusID:4766599}
}

@article{van2023mealpy,
   title={MEALPY: An open-source library for latest meta-heuristic algorithms in Python},
   author={Van Thieu, Nguyen and Mirjalili, Seyedali},
   journal={Journal of Systems Architecture},
   year={2023},
   publisher={Elsevier},
   doi={10.1016/j.sysarc.2023.102871}
}

@misc{zheng2025blackboxbenchcomprehensivebenchmarkblackbox,
      title={BlackboxBench: A Comprehensive Benchmark of Black-box Adversarial Attacks}, 
      author={Meixi Zheng and Xuanchen Yan and Zihao Zhu and Hongrui Chen and Baoyuan Wu},
      year={2025},
      eprint={2312.16979},
      archivePrefix={arXiv},
      primaryClass={cs.CR},
      url={https://arxiv.org/abs/2312.16979}, 
}

\end{document}